*Manuscript accepted for publication in The Proceedings of the 29th National Conference on Selected Issues of Information and Communication Technology (VNICT 2026) - Hanoi, 7-8/11/2026*

# Automated Extraction of Records of Processing Activities (RoPA) Using Hybrid RAG and Locally Deployed Large Language Models

To Duy Hinh[1], Nguyen Le Quoc Anh[1], Phan Van Tri[1] and Khuong Nguyen-An[2,3,*]
[1] Academy of Cryptography Techniques, Ho Chi Minh City Campus, Government Cipher Committee, Vietnam
[2] Faculty of Computer Science and Engineering, Ho Chi Minh City University of Technology (HCMUT), 268 Ly Thuong Kiet Street, Dien Hong Ward, Ho Chi Minh City, Vietnam
[3] Vietnam National University Ho Chi Minh City, Linh Xuan Ward, Ho Chi Minh City, Vietnam
Emails: tdhinhit@gmail.com, haniz.cons@gmail.com, phanvantri@actvn.edu.vn, nakhuong@hcmut.edu.vn

**Abstract - Vietnam's Personal Data Protection Law (Law 91/2025/QH15) and Decree 356/2025/ND-CP, effective January 1, 2026, require organizations to create and maintain Records of Processing Activities (RoPA) under Article 31. Preparing RoPA manually requires considerable staff time, while the strongest large language models (LLMs) run on foreign cloud services, which conflicts with data sovereignty requirements. This study proposes RoPA Manager, a system that extracts RoPA information by combining hybrid retrieval (lexical ranking over a tsvector index, dense vector search, and Reciprocal Rank Fusion, RRF) with locally deployed LLMs. The main contribution is the first Vietnamese RoPA benchmark, with 32 organizations, 77 processing activities, 12 field groups, and 4,338 reference values. Results are reported separately at three evaluation levels. The automated scorer, evaluated on noisy data without calling an LLM, reaches F1 = 0.9493 [0.9436; 0.9548]; this value measures the quality of the scorer, not the accuracy of end-to-end extraction. End-to-end extraction with LLM calls reaches 50.04-55.25% token coverage against the reference labels. Two independent experts reviewed 1,558 reference values (35.9% of the benchmark); no value was rated Wrong, and agreement was 99.68% with PABAK = 0.9936. Value-level precision has not yet been measured. Paired tests on 32 scenarios show no significant difference between the local Qwen3.5-27B-GPTQ-Int4 model on a 24 GB GPU and the cloud-based DeepSeek-V4-Flash (a 0.20 percentage-point difference in favor of DeepSeek, 95% CI [-0.93; +1.32], p = 0.72), while Gemma-4-31B performs significantly worse (p < 0.01).**

## I. INTRODUCTION

Vietnam's Personal Data Protection Law (Law 91/2025/QH15) [2], effective from January 1, 2026, together with Decree 356/2025/ND-CP [1], which provides detailed guidance, requires every organization that processes personal data to create and maintain Records of Processing Activities (RoPA) under Article 31 of Law 91/2025/QH15. Preparing RoPA manually requires strong legal expertise, takes considerable staff time, and is prone to errors, especially in organizations that run dozens of processing activities across many business areas.

Large language models (LLMs) combined with retrieval-augmented generation (RAG) [3] make it possible to automate the extraction of information from business documents into a standard RoPA structure. However, using cloud LLM application programming interfaces (APIs) to process personally identifiable information (PII) conflicts with the principle of data sovereignty. A solution is therefore needed that uses LLMs without sending personal data outside the organization's internal infrastructure.

This paper presents RoPA Manager with three contributions. First, it provides the first benchmark for Vietnamese RoPA extraction, with 32 simulated organizations, 77 processing activities, 12 field groups, and 4,338 reference values, of which 35.9% were reviewed independently by two experts, together with an automated scorer. Second, it introduces a three-level reporting scheme that separates scorer performance, end-to-end coverage, and review of the reference labels. It is intended to avoid mixing three different kinds of metrics, a common problem in published evaluations of RAG systems. Third, it empirically examines the feasibility of local deployment: two open-source models completed all 32 scenarios on a 24 GB GPU, and one of them reached coverage in the same range as the reference cloud model. The model-independent architecture is a design choice that serves the third contribution, not a separate scientific contribution.

RoPA directly supports the confidentiality, integrity, and availability of personal data, while the system itself must defend against LLM-specific threats such as prompt injection and data leakage through cloud models. RoPA Manager therefore builds security principles based on the STRIDE framework and the Zero Trust model into the system from the design stage, as described in Section III.D.

## II. RELATED WORK

Governance, risk, and compliance (GRC) tools such as OneTrust [6] and the ISO/IEC 29134:2023 guidelines [7] support RoPA management but still require a data protection officer (DPO) to enter data manually. NLP-based compliance automation, such as checking the completeness of privacy policies against GDPR requirements [19], has been shown to be feasible but is tied to the European legal framework and to English. Legal information extraction benchmarks such as LexGLUE [20] also cover English only. Vietnamese pretrained models such as PhoBERT [8] have not been applied to the RoPA task. Work on domain-specific RAG [4], [5], [12] shows that retrieval quality and hallucination remain open challenges. A search of Google Scholar, IEEE Xplore, and the ACM Digital Library for 2020-2026, combining "RoPA" or "records of processing activities" with "Vietnamese", "LLM", and "RAG", found no work on automated RoPA extraction for Vietnamese. To the best of the authors' knowledge, this is the gap that this paper addresses.

At the technical level, the Transformer architecture with self-attention [14] is the basis of modern LLMs. LLMs are used instead of rule-based methods because of the nature of the input: the source documents for RoPA are internal procedures, contracts, and impact assessment records. These documents are semi-structured and their layouts differ across organizations, so regular expressions or fixed extraction rules cannot cover them. This is the fundamental difference from tasks with a fixed input schema, where a rule engine is often enough. Hybrid retrieval is chosen because Vietnamese legal terms need exact keyword matching, in the spirit of the probabilistic relevance framework [21], in addition to semantic similarity. RRF [22] merges the two rankings using rank positions only, so it does not depend on the score scale of each retrieval branch. The multilingual-e5-base embedding model [23] is used instead of PhoBERT [8] because the system needs Vietnamese-English support and direct compatibility with pgvector [15].

## III. SYSTEM DESIGN AND METHODS

### *A. Overall architecture*

RoPA Manager uses a layered architecture with four layers: Presentation (single-page web application), Business logic (coordination of the extraction workflow), Artificial Intelligence (model-independent RAG + LLM layer), and Data (relational database and vector database). A key design point is that the AI layer is endpoint-independent, so the system can switch between cloud and local LLMs through configuration alone.

Architecture choices follow two goals: keeping personal data inside the internal infrastructure and minimizing the number of components to operate. The vector database uses pgvector [15] inside PostgreSQL instead of a separate dedicated vector system. The local inference server uses vLLM, whose paged attention reduces GPU memory use and whose OpenAI-compatible interface allows the model provider to be changed through endpoint configuration alone.

### *B. Hybrid retrieval and extraction workflow*

The workflow has six stages (Figure 1): receiving and preprocessing documents (PDF/DOCX); chunking the text and creating embeddings; hybrid retrieval; calling the LLM for extraction; post-processing and validation against the schema; and storing the result for review. Semantic chunking uses a break threshold of 0.62. The sparse branch uses the ts_rank_cd lexical ranking function on a PostgreSQL tsvector index (simple configuration), while the dense branch uses cosine search through an HNSW index [13]. Each branch over-retrieves 30 candidates before fusion. The two branches are merged by a weighted variant of RRF [22] with $k = 60$ and a weight of 0.4 for the lexical branch; both values are defaults and have not been systematically tuned. The top 8 chunks are kept as context, and the evaluated version does not use cross-encoder reranking. The LLM receives a fixed prompt, identified by its SHA-256 checksum, together with the context. The output is constrained to JSON and validated against a Pydantic schema. Figure 1 and Table I report the main parameters for reproducibility.

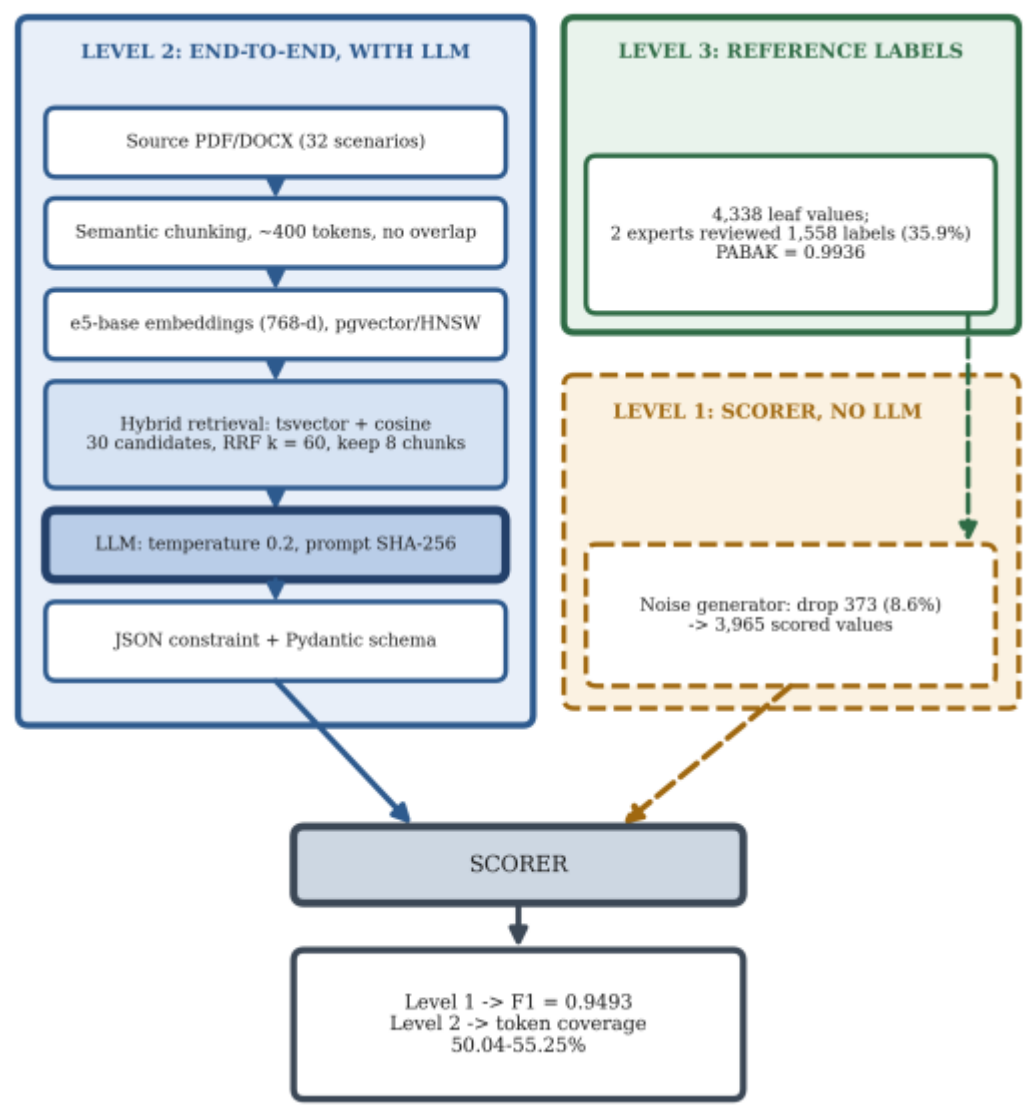

Figure 1. Three evaluation levels. Level 2 runs the full workflow with LLM calls; level 1 follows the dashed branch and passes noise-modified reference labels to the scorer without an LLM; level 3 reviews the reference labels themselves.

Table I. RETRIEVAL CONFIGURATION AND REPRODUCIBILITY PARAMETERS

| Component | Configuration used in experiments |
|---|---|
| Vector index | pgvector / PostgreSQL 15; HNSW m = 16, ef_construction = 64 |
| Generation | temperature = 0.2; max_tokens = 1800 |
| Output constraint | JSON + Pydantic 2.5.3 schema |
| Local inference | vLLM, single 24 GB GPU |
| Platform / seed | Python 3.11, FastAPI 0.109.0; seed 20260705 |

### C. *RoPA field groups*

RoPA information is organized into 12 independent field groups based on Article 31 of Law 91/2025/QH15 [2] and the detailed guidance in Decree 356/2025/ND-CP [1]. The groups are BASIC_INFO; PURPOSES; LEGAL_BASIS; DATA_SUBJECTS for data subjects and data types; RECIPIENTS_TRANSFERS for recipients and cross-border transfers; RETENTION for retention periods and deletion methods; TECHNICAL_MEASURES and STORAGE_SECURITY; CONSENT_DETAILS for collecting and withdrawing consent; RISK_DPIA for impact assessment and risk level; REGULATORY for reporting to the regulator; and DOCUMENTATION. The 12 reporting groups have a many-to-many mapping to 12 query groups in the extraction layer: PURPOSES, DATA_SUBJECTS, and RECIPIENTS_TRANSFERS each combine two queries; STORAGE_SECURITY and TECHNICAL_MEASURES share one query; and DOCUMENTATION has no query of its own but is derived from the description field and the cited sources. The reporting group is the unit of both the reference labels and the measurements.

### D. *Information security design*

Under the OWASP classification for LLM applications [18], prompt injection is the top risk for systems based on large language models. RoPA Manager applies the Zero Trust model of NIST SP 800-207 [17] together with STRIDE threat analysis [16] to identify both traditional attack vectors (spoofing, data tampering, denial of service) and AI-specific threats (prompt injection, data leakage through cloud models, and hallucinated false information). Three defense layers are built directly into the extraction workflow: Pydantic schema validation rejects any output that does not follow the JSON structure; the model's built-in safety mechanism refuses instructions outside the assigned task; and the JSON output constraint greatly limits information leakage through free text. The effectiveness of this design is tested experimentally in Section V.

## IV. EXPERIMENTAL SETUP

### A. *Infrastructure and models*

The system is evaluated with three LLMs: DeepSeek-V4-Flash through a cloud API as the reference, and Qwen3.5-27B-GPTQ-Int4 (4-bit quantized) and Gemma-4-31B (bf16), both deployed locally through vLLM on a single 24 GB GPU. The application layer runs on an AWS EC2 t3.large instance (2 vCPU, 8 GiB RAM, no GPU) because the embedding model runs efficiently on a CPU at this experimental scale. The experiments call the cloud API only with fully synthetic data, so they do not violate the data sovereignty principle; with real data, the system must switch to local mode. All three models use the same source code, the same prompt template (with the same SHA-256 checksum), and the same scorer.

### B. *Benchmark dataset*

The benchmark is built from controlled synthetic data in four steps: selecting seven sectors with heavy RoPA obligations (banking/finance, e-commerce, telecommunications, healthcare, education, information technology/software as a service, and aviation); designing 32 simulated organizations with 1-5 processing activities each; writing unstructured Vietnamese source documents that also contain English terms; and labeling the 12 field groups according to a standardized guide with a conservative rule, checked by a JSON schema validator and logical cross-checks. The documents and reference labels were written manually by the authors and were not generated by an LLM, which avoids the risk of model feedback loops. The difficulty distribution is 21 easy, 33 medium, and 23 hard processing activities. Three measurement units are used in parallel: 924 group entities (77 x 12), 4,338 leaf values in identity mode, and 3,965 leaf values scored in noisy/anonymized mode after 373 fields (8.6%) were removed.

### C. Measurement method

The study separates three evaluation levels because they use different inputs and measure different objects (Figure 1). Level 1 tests the scorer in identity mode and in noisy/anonymized mode; both modes compare the reference labels with versions of them modified in a controlled way, so no LLM is called. Level 2 runs the full workflow with LLM calls on 32 scenarios. The score of each field group is the share of reference-label tokens that are matched, so it is a recall-like measure at the token level. Because the 32 scenarios are paired observations, model comparisons use a paired t-test with a Wilcoxon check. Level 3 reviews the reliability of the reference labels themselves. The 95% confidence intervals are computed by nonparametric bootstrap (B = 10,000, seed = 20260712, N = 3,965); the KPI thresholds in Table II were set by the authors. The level 3 review sample contains 1,558 reference values (35.9% of the benchmark), drawn from 15 of the 32 scenarios and 41 of the 77 processing activities, and covers all 7 sectors and all 12 field groups. Samples were taken in scenario order within each sector block, not by stratified random sampling, so the sampling rate ranges from 33% to 100% across sectors. Two experts scored separate copies independently, without sharing results and without knowing which model produced the labels, on a three-point scale: Correct / Partly correct / Wrong.

### D. Reproducibility

Before each comparison, every component that can affect the result is frozen at a specific version: dataset v2.0, scorer v1.3, source code at git commit c38e18c for the cloud configuration and 045547f for the two local models, and the shared retrieval parameters and prompt template, with integrity checked by SHA-256. Each run records a run ID, provider, quantization method, temperature, seed, and execution timestamp. The source code, the benchmark, and the review records of the two experts are publicly available at https://github.com/tdhinhit/RoPA-demo.

## V. RESULTS AND DISCUSSION

At the scorer level, identity mode gives 1.0 on every metric over 4,338 leaf values, which confirms that the scoring logic has no technical error. In noisy/anonymized mode with 3,965 leaf values, F1 = 0.9493 [0.9436; 0.9548], Precision = 0.9470, and Recall = 0.9517, with confidence intervals narrower than 0.02. Because this mode does not call an LLM, these values only show that the scorer tolerates imperfect (noisy) data; they do not measure extraction performance. Six of the seven level 1 metrics meet their thresholds. The exception is the correct refusal rate, measured on 795 fields whose reference label is null, which reaches only 0.7874 against a threshold of 0.80 (Table II).

Table II. AUTOMATED SCORER RESULTS (NO LLM) AGAINST KPI THRESHOLDS

| **Metric** | **Value (95% CI)** | **Threshold** | **Status** |
|---|---|---|---|
| Level 1 - scorer, noisy mode (N = 3,965, no LLM) | | | |
| F1 | 0.9493 [0.9436; 0.9548] | >= 0.85 | PASS |
| Accuracy | 0.9188 [0.9100; 0.9271] | >= 0.85 | PASS |
| Precision | 0.9470 [0.9390; 0.9547] | >= 0.85 | PASS |
| Recall | 0.9517 [0.9441; 0.9591] | >= 0.80 | PASS |
| MCC [11] | 0.7448 | >= 0.55 | PASS |
| Hallucination (noisy data) | 0.0426 | <= 0.05 | PASS |
| Correct refusal rate (n = 795) | 0.7874 | >= 0.80 | FAIL |

Analysis by the 12 field groups (Table III) shows that PURPOSES, DATA_SUBJECTS, and DOCUMENTATION reach almost perfect F1 scores. RETENTION is the lowest at 0.712, followed by STORAGE_SECURITY at 0.876 and REGULATORY at 0.886; across sectors, F1 stays within a narrow range of 0.939-0.960. The simulated error pattern contains 169 false positives: 55% (93) are values inserted by the noise generator, mainly in RETENTION (46), RISK_DPIA (23), and STORAGE_SECURITY (23), and 45% (76) fall in pairs of groups with overlapping meanings. Of the 153 false negatives, most are in BASIC_INFO (44), LEGAL_BASIS (23), and REGULATORY (18). Because these errors come from artificial noise rather than model output, this pattern is only a predictive hypothesis about the field groups where an LLM is likely to hallucinate or confuse values in operation; testing it on real level 2 output is future work.

Table III. SCORER RESULTS BY 12 FIELD GROUPS (NOISY MODE, NO LLM)

| Field group | TP | FP | FN | P | R | F1 |
|---|---|---|---|---|---|---|
| PURPOSES | 174 | 0 | 0 | 1.000 | 1.000 | 1.000 |
| DATA_SUBJECTS | 566 | 0 | 8 | 1.000 | 0.986 | 0.993 |
| DOCUMENTATION | 200 | 0 | 5 | 1.000 | 0.976 | 0.988 |
| BASIC_INFO | 572 | 0 | 44 | 1.000 | 0.929 | 0.963 |
| TECHNICAL_MEASURES | 303 | 24 | 0 | 0.927 | 1.000 | 0.962 |
| RECIPIENTS_TRANSFERS | 202 | 18 | 0 | 0.918 | 1.000 | 0.957 |
| LEGAL_BASIS | 235 | 0 | 23 | 1.000 | 0.911 | 0.953 |
| CONSENT_DETAILS | 209 | 18 | 13 | 0.921 | 0.941 | 0.931 |
| RISK_DPIA | 214 | 23 | 17 | 0.903 | 0.926 | 0.915 |
| REGULATORY | 136 | 17 | 18 | 0.889 | 0.883 | 0.886 |
| STORAGE_SECURITY | 138 | 23 | 16 | 0.857 | 0.896 | 0.876 |
| RETENTION | 68 | 46 | 9 | 0.597 | 0.883 | 0.712 |

At the end-to-end extraction level, Table IV compares the three models on the same dataset, source code, and prompt template. All three complete 32/32 scenarios, which confirms that the model-independent architecture works as designed. Token coverage is 55.25% for DeepSeek-V4-Flash, 55.06% for the local Qwen3.5-27B, and 50.04% for Gemma-4-31B. Because the 32 scenarios are paired observations, the study uses a paired t-test with a Wilcoxon check. The difference between DeepSeek and Qwen is only +0.20 percentage points (95% CI [-0.93; +1.32], $t(31) = 0.36$, $p = 0.72$; Wilcoxon $p = 0.61$). No significant difference is found, and the confidence interval also excludes differences larger than about 1.3 percentage points, so this result is evidence of equivalence within about 1.3 percentage points, not only a lack of evidence for a difference. In contrast, Gemma is 5.21 percentage points below DeepSeek ([+2.03; +8.40], $p = 0.0022$) and 5.02 percentage points below Qwen ([+2.10; +7.94], $p = 0.0014$). The gap lies mainly in LEGAL_BASIS, where Gemma reaches only 41.6-56.1% coverage in four sectors, compared with 83.5-100% for the other two models. The tests rely on the variation across scenarios within a single run.

Table IV. COMPARISON OF THREE MODELS ON THE SAME DATASET AND PROMPT

| **Criterion** | **DeepSeek-V4-Flash** | **Qwen3.5-27B-Int4** | **Gemma-4-31B** |
|---|---|---|---|
| Deployment | Cloud API | Local vLLM | Local vLLM |
| Parameters | N/A (black box) | 27B (4-bit) | 31B (bf16) |
| Data location (design) | PII leaves infrastructure | PII internal | PII internal |
| Completed scenarios | 32/32 | 32/32 | 32/32 |
| Coverage (1 run) | 55.25% | 55.06% | 50.04% |
| Standard deviation | 4.43 | 4.63 | 9.30 |

Chunking quality directly affects the retrieval stage. The operational workflow uses semantic chunking (Figure 1), with an average of 42 chunks per document. A separate control run uses recursive character-based chunking, which produces only 6.9 chunks per document on average. Across 32 documents, both methods keep 100% of chunks within the 512-token limit of the embedding model and do not break sentences; the semantic chunks also have high internal cohesion (cosine 0.786-0.832). One consequence deserves attention: with the policy of keeping 8 chunks as context, the current semantic setting puts only about 19% of each document into the context of each query, while the recursive setting would include almost the whole document. This is a significant competing hypothesis for the level 2 token coverage of 50.04-55.25%, and an ablation study is needed to test it.

For the retrieval stage, the study has not measured Recall@K, MRR, or NDCG, because no query-chunk relevance benchmark has been built yet. Without this separate measurement, the paper does not attribute the low coverage to either the retrieval stage or the generation stage.

Regarding the reliability of the reference labels, no value was rated Wrong. The Correct rates of the two experts are 99.81% and 99.74%, and inter-expert agreement is 99.68% (1,553/1,558), with 5 disagreements. All five concern the same issue: whether Articles 13 and 14 of Law 91/2025/QH15 must be cited in cases of automated credit scoring, exam proctoring by face recognition, and AI-based staff assessment. This shows that the hardest part of the task is legal reasoning rather than text extraction. These cases are also rated hard, so extraction difficulty and labeling difficulty tend to go together. Combined with the high share of hard activities in the review sample (46.3% versus 29.9% in the full benchmark), this indicates that the sample is not biased toward easy cases. Because 99.7% of the labels fall in the same class, raw Cohen's kappa is only 0.28, which is the known prevalence paradox. The study therefore uses PABAK = 0.9936 [10] as the main agreement measure and reports raw kappa for transparency. Two limits should be noted: a high PABAK only shows that the two experts are consistent with each other, and both experts come from the same business ecosystem.

For information security, the study uses two types of evaluation. First, a review of the operating configuration checks 12 HTTP response-header settings of the reverse proxy and gives 4 PASS, 6 WARN, and 2 FAIL results. The two FAIL items are a missing HSTS header and an API endpoint that does not require authentication; both must be fixed before operation. Second, an attack-resistance test shows that the system blocks all 8 prompt-injection scenarios: direct instruction injection, administrator role-play, forced system-prompt disclosure, indirect injection through a document field, personal-data extraction, JSON-schema override, multi-turn context poisoning, and Unicode encoding. This result comes from the three defense layers described in Section III.D. With n = 8 and a single run, it is only a proof of concept and does not support a strong claim about attack resistance.

The evaluation framework has three levels with different scopes. Level 1 evaluates the scorer, level 2 measures token coverage against the reference labels in a recall-like way, and level 3 evaluates the reliability of the reference labels. None of the levels measures value-level precision on real model output, that is, the share of generated values that are correct or the share of blank fields that are correctly left blank; level 1 measures these two quantities only for the scorer, without LLM calls. All conclusions in this paper are therefore limited to technical feasibility and coverage.

## VI. CONCLUSION AND FUTURE WORK

This study sets a technical baseline for applying LLM + RAG to automate compliance with Vietnam's Personal Data Protection Law, and contributes the first Vietnamese RoPA benchmark and a three-level reporting scheme. As a cost reference, a 24 GB GPU costs about VND 50-65 million, a full-time DPO about VND 180-300 million per year, and a commercial GRC package about USD 10,000-50,000 per year. In order of priority, future work includes measuring value-level content accuracy on real drafts, with stratified sampling, two independent reviewers, three labels (correct / wrong / extra generated value), and precision and recall reported by field group; repeating the multi-model experiments at least three times; building a relevance benchmark to measure Recall@K, MRR, and NDCG, and running an ablation study for hybrid retrieval [5]; and measuring the calibration of model-generated confidence scores on real outputs.

The main limitations are: (1) all data are synthetic and have not been tested on real RoPA records; (2) value-level precision has not been measured; (3) level 1 metrics evaluate the scorer without LLM calls, so the error pattern by field group is only a predictive hypothesis; (4) the paired tests use a single run; (5) no ablation study has measured the separate contribution of RRF, and Recall@K, MRR, and NDCG have not been measured; (6) the reference-label review covers 35.9% of the benchmark, samples scenarios in order rather than at random, and uses n = 2 experts; and (7) the calibration of model-generated confidence scores on real output has not been evaluated [9]. These limitations define the contribution as a technical baseline; the study does not claim full operational legal compliance.

## ACKNOWLEDGMENTS

The authors thank two independent compliance and legal experts for their single-blind review of 1,558 reference values, which provides a reliable basis for all measurements in this study. The study was self-funded by the authors and received no funding from any organization or company. The authors have no commercial interest in the described system and no financial relationship with the model providers or compliance management tools mentioned in this paper, and declare no conflict of interest. Khuong Nguyen-An thanks Ho Chi Minh City University of Technology (HCMUT), VNU-HCM, for supporting this research.

# Tự động hóa trích xuất Hồ sơ hoạt động xử lý dữ liệu cá nhân (RoPA) bằng RAG lai và Mô hình ngôn ngữ lớn triển khai tại chỗ

Tô Duy Hinh[1], Nguyễn Lê Quốc Anh[1], Phan Văn Trị[1] và Nguyễn An Khương[2,3,(✉)]

[1] Học viện Kỹ thuật Mật mã, Cơ sở TP. Hồ Chí Minh, Ban Cơ yếu Chính phủ, Việt Nam
[2] Khoa Khoa học và Kỹ thuật Máy tính, Trường Đại học Bách khoa (HCMUT),
268 Lý Thường Kiệt, Phường Diên Hồng, TP. Hồ Chí Minh, Việt Nam
[3] Đại học Quốc gia TP. Hồ Chí Minh, Phường Linh Xuân, TP. Hồ Chí Minh, Việt Nam
Emails: tdhinhit@gmail.com, haniz.cons@gmail.com, phanvantri@actvn.edu.vn, nakhuong@hcmut.edu.vn

**Abstract - Luật Bảo vệ dữ liệu cá nhân (Luật 91/2025/QH15) và Nghị định 356/2025/NĐ-CP (hiệu lực 01/01/2026) buộc tổ chức lập và duy trì Hồ sơ hoạt động xử lý dữ liệu cá nhân (Records of Processing Activities, RoPA) theo Điều 31. Việc lập RoPA thủ công tốn nhiều nhân lực, trong khi các Mô hình ngôn ngữ lớn (Large Language Model, LLM) mạnh nhất vận hành trên đám mây nước ngoài, mâu thuẫn với yêu cầu chủ quyền dữ liệu. Nghiên cứu đề xuất RoPA Manager, hệ thống trích xuất thông tin RoPA bằng truy xuất lai kết hợp xếp hạng từ vựng trên chỉ mục tsvector, vector mật độ cao và dung hợp thứ hạng tương hỗ (Reciprocal Rank Fusion, RRF) với LLM triển khai tại chỗ. Đóng góp chính là bộ kiểm chuẩn RoPA tiếng Việt đầu tiên gồm 32 tổ chức, 77 hoạt động xử lý, 12 nhóm trường và 4338 giá trị tham chiếu. Kết quả được báo cáo tách bạch theo ba tầng đo. Bộ chấm điểm tự động, đo trên dữ liệu làm nhiễu và không gọi LLM, đạt F1 = 0.9493 [0.9436; 0.9548]; đây là chất lượng của bộ chấm điểm, không phải độ chính xác của hệ trích xuất đầu-cuối. Trích xuất đầu-cuối có gọi LLM cho độ phủ token so với nhãn chuẩn 50.04-55.25%. Hai chuyên gia độc lập thẩm định 1558 giá trị nhãn chuẩn (35.9% bộ kiểm chuẩn), không giá trị nào sai, đồng thuận 99.68% với PABAK = 0.9936. Nghiên cứu chưa đo độ chính xác ở cấp giá trị theo hướng precision. Kiểm định ghép cặp trên 32 kịch bản cho thấy mô hình tại chỗ Qwen3.5-27B-GPTQ-Int4 không khác biệt so với mô hình đám mây DeepSeek-V4-Flash (chênh lệch 0.20 điểm nghiêng về DeepSeek, CI 95% [-0.93; +1.32], p = 0.72) trên GPU 24 GB, còn Gemma-4-31B thấp hơn có ý nghĩa (p < 0.01).**

## I. GIỚI THIỆU

Luật Bảo vệ dữ liệu cá nhân (Luật 91/2025/QH15) [2] có hiệu lực từ 01/01/2026, cùng Nghị định 356/2025/NĐ-CP [1] hướng dẫn chi tiết, yêu cầu mọi tổ chức xử lý dữ liệu cá nhân lập và duy trì Hồ sơ hoạt động xử lý (RoPA) theo Điều 31 Luật 91/2025/QH15. Việc lập RoPA thủ công đòi hỏi chuyên môn pháp lý sâu, tốn nhiều nhân lực và dễ sai sót, nhất là với tổ chức vận hành hàng chục hoạt động xử lý trải rộng nhiều lĩnh vực.

Mô hình ngôn ngữ lớn (LLM) kết hợp kiến trúc Tạo sinh tăng cường bằng truy xuất (RAG) [3] cho phép tự động hóa trích xuất thông tin từ tài liệu nghiệp vụ sang cấu trúc RoPA chuẩn. Tuy nhiên, dùng LLM qua giao diện lập trình ứng dụng (API) đám mây để xử lý dữ liệu nhận dạng cá nhân (PII) mâu thuẫn với nguyên tắc chủ quyền dữ liệu. Do đó cần giải pháp tận dụng LLM mà dữ liệu cá nhân không rời khỏi hạ tầng nội bộ.

Bài báo trình bày RoPA Manager với ba đóng góp. Thứ nhất là bộ kiểm chuẩn đầu tiên cho trích xuất RoPA tiếng Việt, gồm 32 tổ chức giả lập, 77 hoạt động xử lý, 12 nhóm trường và 4338 giá trị tham chiếu, trong đó 35.9% đã qua thẩm định độc lập của hai chuyên gia, kèm bộ chấm điểm tự động. Thứ hai là nguyên tắc báo cáo ba tầng, tách bạch bộ chấm điểm, độ phủ đầu-cuối và thẩm định nhãn chuẩn, nhằm tránh việc gộp lẫn ba loại chỉ số vốn phổ biến khi công bố hệ thống RAG. Thứ ba là khảo sát thực nghiệm tính khả thi của triển khai tại chỗ, trong đó hai mô hình mã nguồn mở chạy trọn 32/32 kịch bản trên GPU 24 GB và một mô hình đạt độ phủ cùng khoảng với mô hình đám mây tham chiếu. Kiến trúc độc lập mô hình là lựa chọn thiết kế phục vụ đóng góp thứ ba, không phải đóng góp khoa học độc lập.

RoPA phục vụ trực tiếp bộ ba Bảo mật, Toàn vẹn và Sẵn sàng của dữ liệu cá nhân, trong khi hệ thống phải phòng chống các mối đe dọa đặc thù của LLM như tiêm nhiễm câu lệnh và rò rỉ dữ liệu qua mô hình đám mây. Do đó, RoPA Manager tích hợp ngay từ khâu thiết kế các nguyên lý bảo mật theo khung STRIDE và mô hình không tin cậy mặc định (Zero Trust), trình bày tại Mục III.D.

## II. NGHIÊN CỨU LIÊN QUAN

Nhóm công cụ quản trị tuân thủ (GRC) như OneTrust [6] và hướng dẫn ISO/IEC 29134:2023 [7] hỗ trợ quy trình quản lý RoPA nhưng vẫn yêu cầu chuyên viên bảo vệ dữ liệu (DPO) nhập liệu thủ công. Nhóm tự động hóa tuân thủ bằng NLP, tiêu biểu là kiểm tra tính đầy đủ của chính sách quyền riêng tư đối chiếu GDPR [19], khả thi nhưng gắn với khung pháp lý châu Âu và tiếng Anh; các bộ kiểm chuẩn trích xuất thông tin pháp lý như LexGLUE [20] cũng chỉ phục vụ tiếng Anh. Mô hình tiền huấn luyện tiếng Việt như PhoBERT [8] chưa được áp dụng cho bài toán RoPA. Nhóm RAG cho miền chuyên ngành [4], [5], [12] cho thấy chất lượng truy xuất và ảo giác vẫn là thách thức mở. Khảo sát trên Google Scholar, IEEE Xplore và ACM Digital Library (RoPA / records of processing activities kết hợp Vietnamese, LLM, RAG; 2020-2026) không tìm thấy công trình nào về trích xuất RoPA tự động cho tiếng Việt; theo hiểu biết của nhóm tác giả, đây là khoảng trống bài báo hướng tới.

Về nền tảng kỹ thuật, kiến trúc Transformer với cơ chế tự chú ý [14] là cơ sở của các LLM hiện đại. Việc dùng LLM thay cho phương pháp dựa trên quy tắc xuất phát từ đặc thù đầu vào: tài liệu nguồn phục vụ RoPA là quy trình nội bộ, hợp đồng và hồ sơ đánh giá tác động, ở dạng bán phi cấu trúc và không đồng nhất về bố cục giữa các tổ chức, nên biểu thức chính quy hay luật trích xuất cố định không bao phủ được. Đây là khác biệt cơ bản so với các bài toán có lược đồ đầu vào cố định, nơi máy luật thường đủ. Truy xuất lai được chọn vì thuật ngữ pháp lý tiếng Việt đòi hỏi khớp chính xác theo từ khóa, theo tinh thần của khung xếp hạng xác suất [21], song song với tương đồng ngữ nghĩa, và RRF [22] hợp nhất hai bảng xếp hạng chỉ dựa trên thứ hạng nên không phụ thuộc thang điểm của từng luồng. Mô hình nhúng đa ngôn ngữ multilingual-e5-base [23] được chọn thay vì PhoBERT [8] vì cần hỗ trợ song ngữ Việt-Anh và tương thích trực tiếp với pgvector [15].

## III. THIẾT KẾ HỆ THỐNG VÀ PHƯƠNG PHÁP

### *A. Kiến trúc tổng thể*

RoPA Manager theo kiến trúc phân lớp gồm bốn lớp: Trình bày (ứng dụng web đơn trang), Nghiệp vụ (điều phối quy trình trích xuất), Trí tuệ nhân tạo (lớp RAG + LLM độc lập mô hình) và Dữ liệu (cơ sở dữ liệu quan hệ và cơ sở dữ liệu vector). Điểm đáng lưu ý của thiết kế là lớp trí tuệ nhân tạo độc lập điểm cuối, cho phép chuyển đổi giữa LLM đám mây và LLM tại chỗ chỉ qua cấu hình.

Các quyết định kiến trúc theo tiêu chí giữ dữ liệu cá nhân trong hạ tầng nội bộ và tối thiểu số thành phần vận hành: cơ sở dữ liệu vector dùng pgvector [15] tích hợp trong PostgreSQL thay vì một hệ vector chuyên dụng riêng; máy chủ suy luận tại chỗ dùng vLLM nhờ paged attention giúp giảm bộ nhớ GPU và cung cấp giao diện tương thích OpenAI, cho phép chuyển đổi nhà cung cấp mô hình chỉ bằng cấu hình điểm cuối.

### *B. Quy trình truy xuất lai và trích xuất*

Quy trình gồm sáu giai đoạn (Hình 1): tiếp nhận và tiền xử lý tài liệu (PDF/DOCX); phân đoạn và sinh vector nhúng; truy xuất lai; gọi LLM trích xuất; hậu xử lý và kiểm tra hợp lệ theo lược đồ; lưu trữ để rà soát. Phân đoạn ngữ nghĩa dùng ngưỡng ngắt 0.62. Luồng thưa dùng hàm xếp hạng từ vựng ts_rank_cd trên chỉ mục tsvector của PostgreSQL (cấu hình simple); luồng mật độ cao dùng tìm kiếm cosine qua chỉ mục HNSW [13]. Mỗi luồng lấy dư 30 ứng viên trước khi hợp nhất. Hai luồng được hợp nhất bằng biến thể có trọng số của RRF [22] với hằng số $k = 60$ và trọng số 0.4 cho nhánh từ vựng, hai giá trị lấy theo cấu hình mặc định và chưa qua tối ưu hóa có hệ thống, rồi giữ 8 đoạn xếp hạng cao nhất làm ngữ cảnh; phiên bản được đo không bật bước xếp hạng lại bằng bộ mã hóa chéo. LLM nhận câu lệnh cố định (checksum SHA-256) kèm ngữ cảnh; đầu ra bị ràng buộc JSON và xác thực lược đồ Pydantic. Hình 1 và Bảng I công bố các tham số chính để bảo đảm khả năng tái lập.

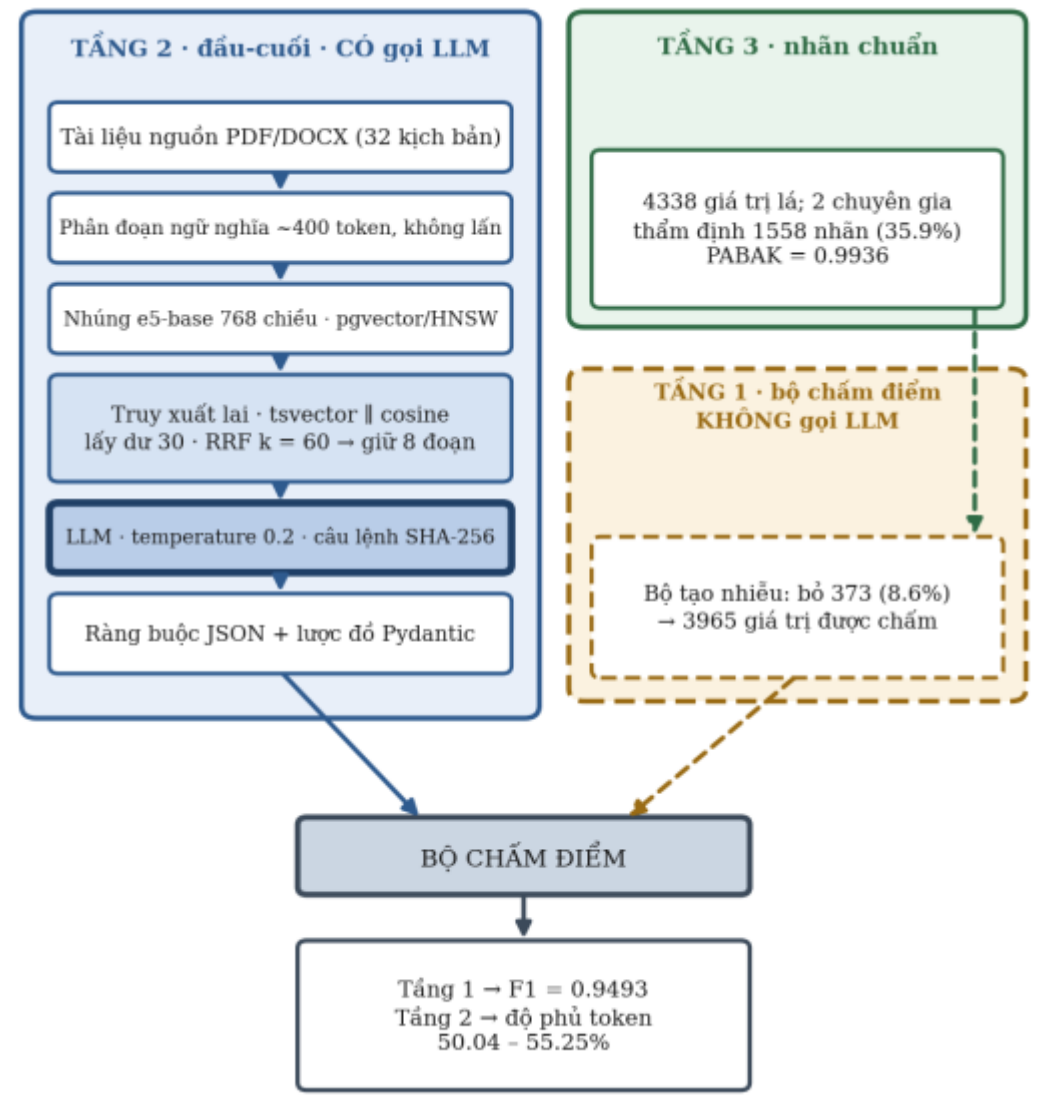

Hình 1. Ba tầng đo của khung đánh giá. Tầng 2 chạy trọn quy trình có gọi LLM; tầng 1 theo nhánh nét đứt, đối chiếu nhãn chuẩn đã làm nhiễu với bộ chấm điểm mà không qua LLM; tầng 3 thẩm định chính nhãn chuẩn.

Bảng I. CẤU HÌNH TRUY XUẤT VÀ THAM SỐ TÁI LẬP

| **Thành phần** | **Cấu hình sử dụng trong thực nghiệm** |
|---|---|
| Chỉ mục vector | pgvector / PostgreSQL 15; HNSW m = 16, ef_construction = 64 |
| Sinh nội dung | temperature = 0.2; max_tokens = 1800 |
| Ràng buộc đầu ra | JSON + lược đồ Pydantic 2.5.3 |
| Suy luận tại chỗ | vLLM, GPU đơn 24 GB VRAM |
| Nền tảng / seed | Python 3.11, FastAPI 0.109.0; seed 20260705 |

### C. *Nhóm trường RoPA*

Thông tin RoPA được tổ chức thành 12 nhóm trường độc lập bám sát Điều 31 Luật 91/2025/QH15 [2] và hướng dẫn chi tiết tại Nghị định 356/2025/NĐ-CP [1], lần lượt là BASIC_INFO, PURPOSES, LEGAL_BASIS, rồi DATA_SUBJECTS cho chủ thể và loại dữ liệu, RECIPIENTS_TRANSFERS cho bên nhận và chuyển dữ liệu xuyên biên giới, RETENTION cho thời hạn lưu trữ và phương thức xóa, hai nhóm TECHNICAL_MEASURES và STORAGE_SECURITY, rồi CONSENT_DETAILS cho việc thu thập và rút lại đồng ý, RISK_DPIA cho đánh giá tác động và mức rủi ro, REGULATORY cho việc báo cáo cơ quan quản lý, và sau cùng là nhóm DOCUMENTATION. Mười hai nhóm báo cáo ánh xạ nhiều-nhiều sang mười hai nhóm truy vấn của lớp trích xuất: PURPOSES, DATA_SUBJECTS và RECIPIENTS_TRANSFERS mỗi nhóm gộp từ hai truy vấn; hai nhóm STORAGE_SECURITY và TECHNICAL_MEASURES dùng chung một truy vấn; riêng DOCUMENTATION không có truy vấn mà suy ra từ trường mô tả và nguồn trích dẫn. Nhóm báo cáo là đơn vị của nhãn chuẩn và của đo lường.

### D. *Thiết kế an toàn thông tin*

Theo phân loại của OWASP dành cho ứng dụng LLM [18], tiêm nhiễm câu lệnh (prompt injection) là rủi ro hàng đầu đối với hệ thống dựa trên mô hình ngôn ngữ lớn. RoPA Manager áp dụng mô hình không tin cậy mặc định (Zero Trust) theo hướng dẫn NIST SP 800-207 [17], kết hợp phân tích mối đe dọa theo khung STRIDE [16] để nhận diện đồng thời các vector tấn công truyền thống (giả mạo, can thiệp dữ liệu, từ chối dịch vụ) và đặc thù AI (tiêm nhiễm câu lệnh, rò rỉ dữ liệu qua mô hình đám mây, ảo giác sinh thông tin sai lệch). Ba lớp phòng thủ được tích hợp trực tiếp vào luồng trích xuất: xác thực lược đồ Pydantic loại bỏ mọi đầu ra không đúng cấu trúc JSON, cơ chế an toàn nội tại của mô hình từ chối thực thi chỉ thị nằm ngoài phạm vi được giao, và ràng buộc đầu ra JSON hạn chế đáng kể khả năng rò rỉ thông tin qua văn bản tự do. Hiệu quả của thiết kế này được kiểm chứng thực nghiệm tại Mục V.

## IV. THIẾT LẬP THỰC NGHIỆM

### A. *Hạ tầng và mô hình*

Hệ thống được đánh giá trên ba LLM: DeepSeek-V4-Flash (API đám mây) làm mốc so sánh, Qwen3.5-27B-GPTQ-Int4 (lượng tử hóa 4-bit) và Gemma-4-31B (bf16) triển khai tại chỗ qua vLLM trên GPU đơn 24 GB VRAM. Tầng ứng dụng chạy trên AWS EC2 t3.large (2 vCPU, 8 GiB RAM, không GPU) vì mô hình nhúng vận hành hiệu quả trên CPU ở quy mô thử nghiệm. Giai đoạn thực nghiệm gọi API đám mây với dữ liệu hoàn toàn tổng hợp nên không vi phạm nguyên tắc chủ quyền dữ liệu; khi vận hành với dữ liệu thật, hệ thống bắt buộc chuyển sang chế độ tại chỗ. Cả ba mô hình dùng chung mã nguồn, mẫu câu lệnh (cùng checksum SHA-256) và bộ chấm điểm.

### B. *Bộ dữ liệu kiểm chuẩn*

Bộ dữ liệu được sinh tổng hợp có kiểm soát qua bốn bước: chọn 7 lĩnh vực có nghĩa vụ RoPA cao (ngân hàng/tài chính, thương mại điện tử, viễn thông, y tế, giáo dục, công nghệ thông tin/phần mềm dạng dịch vụ, hàng không); thiết kế 32 tổ chức giả lập, mỗi tổ chức 1-5 hoạt động xử lý; soạn tài liệu nguồn phi cấu trúc tiếng Việt xen thuật ngữ tiếng Anh; gán nhãn 12 nhóm trường theo hướng dẫn chuẩn hóa với nguyên tắc bảo thủ, kiểm tra bằng bộ xác thực lược đồ JSON và kiểm tra logic chéo. Tài liệu và nhãn chuẩn do nhóm tác giả soạn thủ công, không sinh bằng LLM, nhằm loại trừ rủi ro vòng lặp mô hình. Phân bố độ khó gồm 21 dễ, 33 trung bình và 23 khó. Ba đơn vị đo lường dùng song song: 924 thực thể nhóm (77 × 12); 4338 giá trị lá ở chế độ đồng nhất; và 3965 giá trị lá được chấm ở chế độ làm nhiễu/ẩn danh, sau khi lược bỏ 373 trường (8.6%).

### C. *Phương pháp đo lường*

Nghiên cứu tách bạch ba tầng đo vì chúng có đầu vào và đối tượng khác nhau (Hình 1). Tầng 1 kiểm tra bộ chấm điểm qua chế độ đồng nhất và chế độ làm nhiễu/ẩn danh; cả hai đối chiếu nhãn chuẩn với phiên bản đã biến đổi có kiểm soát nên không gọi LLM. Tầng 2 chạy trọn quy trình có gọi LLM trên 32 kịch bản; điểm mỗi nhóm trường được tính bằng tỉ lệ token khớp với nhãn chuẩn, tức một chỉ số dạng recall ở cấp token. Vì 32 kịch bản là quan sát ghép cặp, so sánh giữa các mô hình dùng kiểm định t ghép cặp kèm đối chứng Wilcoxon. Tầng 3 thẩm định độ tin cậy của chính nhãn chuẩn. Khoảng tin cậy 95% tính bằng bootstrap phi tham số (B = 10000, seed = 20260712, N = 3965); ngưỡng KPI tại Bảng II do nhóm nghiên cứu tự đặt. Mẫu thẩm định ở tầng 3 gồm 1558 giá trị nhãn chuẩn (35.9% bộ kiểm chuẩn), thuộc 15/32 kịch bản và 41/77 hoạt động xử lý, phủ đủ 7/7 lĩnh vực và 12/12 nhóm trường. Mẫu được lấy theo thứ tự kịch bản trong mỗi khối lĩnh vực, không áp dụng thiết kế phân tầng ngẫu nhiên, nên tỉ lệ lấy mẫu dao động 33-100% giữa các lĩnh vực. Hai chuyên gia chấm độc lập trên hai bản sao tách rời, không trao đổi kết quả và không biết nhãn sinh từ mô hình nào, theo thang ba mức Đúng / Đúng một phần / Sai.

### *D. Khả năng tái lập*

Trước mỗi lượt so sánh, mọi thành phần ảnh hưởng kết quả được đóng băng phiên bản: bộ dữ liệu (v2.0), bộ chấm điểm (v1.3), mã nguồn (git commit c38e18c cho cấu hình đám mây và 045547f cho hai mô hình tại chỗ), tham số truy xuất và mẫu câu lệnh dùng chung, kiểm soát toàn vẹn bằng SHA-256. Mỗi lượt chạy ghi nhận mã lượt chạy, nhà cung cấp, phương pháp lượng tử hóa, temperature, seed và thời điểm thực thi. Mã nguồn, bộ kiểm chuẩn và hồ sơ thẩm định của hai chuyên gia được công khai tại địa chỉ https://github.com/tdhinhit/RoPA-demo.

## V. KẾT QUẢ VÀ THẢO LUẬN

Ở tầng bộ chấm điểm, chế độ đồng nhất đạt mọi chỉ số bằng 1.0 trên 4338 giá trị lá, xác nhận logic chấm điểm không có lỗi kỹ thuật. Trên chế độ làm nhiễu/ẩn danh (3965 giá trị lá), F1 = 0.9493 [0.9436; 0.9548], Precision = 0.9470 và Recall = 0.9517, với biên khoảng tin cậy hẹp dưới 0.02. Vì chế độ này không gọi LLM, các con số trên chỉ chứng minh bộ chấm điểm chịu được dữ liệu khiếm khuyết, không phải hiệu năng trích xuất. Sáu trong bảy chỉ số tầng 1 đạt ngưỡng; riêng độ chính xác từ chối trả lời, đo trên 795 trường có nhãn chuẩn null, chỉ đạt 0.7874 so với ngưỡng 0.80 (Bảng II).

Bảng II. KẾT QUẢ BỘ CHẤM ĐIỂM TỰ ĐỘNG (KHÔNG GỌI LLM) SO VỚI NGƯỠNG KPI

| Chỉ số | Giá trị (CI 95%) | Ngưỡng | Trạng thái |
|---|---|---|---|
| **Tầng 1 - bộ chấm điểm, chế độ làm nhiễu (N = 3965, không gọi LLM)** | | | |
| F1 | 0.9493 [0.9436; 0.9548] | ≥ 0.85 | PASS |
| Accuracy | 0.9188 [0.9100; 0.9271] | ≥ 0.85 | PASS |
| Precision | 0.9470 [0.9390; 0.9547] | ≥ 0.85 | PASS |
| Recall | 0.9517 [0.9441; 0.9591] | ≥ 0.80 | PASS |
| MCC [11] | 0.7448 | ≥ 0.55 | PASS |
| Ảo giác (dữ liệu nhiễu) | 0.0426 | ≤ 0.05 | PASS |
| Từ chối trả lời (n = 795) | 0.7874 | ≥ 0.80 | FAIL |

Phân tích theo 12 nhóm trường (Bảng III) cho thấy PURPOSES, DATA_SUBJECTS và DOCUMENTATION đạt F1 gần tuyệt đối, RETENTION thấp nhất với 0.712, tiếp theo REGULATORY 0.886 và STORAGE_SECURITY 0.876; theo lĩnh vực, F1 dao động hẹp 0.939-0.960. Cơ cấu sai số trong chế độ mô phỏng gồm 169 dương tính giả, trong đó 55% (93) là giá trị được bộ tạo nhiễu chèn thêm, tập trung ở RETENTION (46), RISK_DPIA (23) và STORAGE_SECURITY (23), còn 45% (76) rơi vào các cặp nhóm chồng lấn ngữ nghĩa. Trong 153 âm tính giả, phần lớn thuộc về ba nhóm BASIC_INFO (44), LEGAL_BASIS (23) và REGULATORY (18). Vì đây là nhiễu nhân tạo chứ không phải đầu ra mô hình, cơ cấu trên chỉ là giả thuyết dự báo về những nhóm trường mà LLM nhiều khả năng ảo giác hoặc nhầm lẫn khi vận hành; kiểm chứng trên đầu ra thật của tầng 2 là công việc tiếp theo.

Bảng III. KẾT QUẢ BỘ CHẤM ĐIỂM THEO 12 NHÓM TRƯỜNG (CHẾ ĐỘ LÀM NHIỄU, KHÔNG GỌI LLM)

| Nhóm trường | TP | FP | FN | P | R | F1 |
|---|---|---|---|---|---|---|
| PURPOSES | 174 | 0 | 0 | 1.000 | 1.000 | 1.000 |
| DATA_SUBJECTS | 566 | 0 | 8 | 1.000 | 0.986 | 0.993 |
| DOCUMENTATION | 200 | 0 | 5 | 1.000 | 0.976 | 0.988 |
| BASIC_INFO | 572 | 0 | 44 | 1.000 | 0.929 | 0.963 |
| TECHNICAL_MEASURES | 303 | 24 | 0 | 0.927 | 1.000 | 0.962 |
| RECIPIENTS_TRANSFERS | 202 | 18 | 0 | 0.918 | 1.000 | 0.957 |
| LEGAL_BASIS | 235 | 0 | 23 | 1.000 | 0.911 | 0.953 |
| CONSENT_DETAILS | 209 | 18 | 13 | 0.921 | 0.941 | 0.931 |
| RISK_DPIA | 214 | 23 | 17 | 0.903 | 0.926 | 0.915 |
| REGULATORY | 136 | 17 | 18 | 0.889 | 0.883 | 0.886 |
| STORAGE_SECURITY | 138 | 23 | 16 | 0.857 | 0.896 | 0.876 |
| RETENTION | 68 | 46 | 9 | 0.597 | 0.883 | 0.712 |

Ở tầng trích xuất đầu-cuối, Bảng IV so sánh ba mô hình trên cùng bộ dữ liệu, mã nguồn và mẫu câu lệnh. Cả ba hoàn thành 32/32 kịch bản, xác nhận kiến trúc độc lập mô hình vận hành đúng thiết kế. Độ phủ token đạt 55.25% với DeepSeek-V4-Flash, 55.06% với Qwen3.5-27B tại chỗ và 50.04% với Gemma-4-31B. Vì 32 kịch bản là quan sát ghép cặp, nghiên cứu áp dụng kiểm định t ghép cặp kèm đối chứng Wilcoxon. Giữa DeepSeek và Qwen, chênh lệch chỉ +0.20 điểm với khoảng tin cậy 95% [-0.93; +1.32], $t(31) = 0.36$, $p = 0.72$ và Wilcoxon $p = 0.61$. Không phát hiện khác biệt, đồng thời khoảng tin cậy loại trừ mọi chênh lệch lớn hơn khoảng 1.3 điểm, nên đây là bằng chứng ủng hộ tính tương đương chứ không chỉ là thiếu bằng chứng bác bỏ. Ngược lại, Gemma thấp hơn DeepSeek 5.21 điểm [+2.03; +8.40] với p = 0.0022 và thấp hơn Qwen 5.02 điểm [+2.10; +7.94] với p = 0.0014; khoảng cách tập trung tại LEGAL_BASIS, nơi mô hình chỉ đạt 41.6-56.1% độ phủ ở bốn lĩnh vực so với 83.5-100% của hai mô hình còn lại. Kiểm định dựa vào biến thiên giữa các kịch bản trong một lượt chạy.

Bảng IV. SO SÁNH BA MÔ HÌNH TRÊN CÙNG BỘ DỮ LIỆU VÀ CÂU LỆNH

| Tiêu chí | DeepSeek-V4-Flash | Qwen3.5-27B-Int4 | Gemma-4-31B |
|---|---|---|---|
| Triển khai | API đám mây | vLLM tại chỗ | vLLM tại chỗ |
| Tham số | N/A (hộp đen) | 27B (4-bit) | 31B (bf16) |
| Lưu trú dữ liệu (thiết kế) | PII rời hạ tầng | PII nội bộ | PII nội bộ |
| Kịch bản hoàn thành | 32/32 | 32/32 | 32/32 |
| Độ phủ (1 lượt chạy) | 55.25% | 55.06% | 50.04% |
| Độ lệch chuẩn | 4.43 | 4.63 | 9.30 |

Chất lượng phân đoạn ảnh hưởng trực tiếp đến tầng truy xuất. Quy trình vận hành dùng phân đoạn ngữ nghĩa (Hình 1), trung bình 42 đoạn/tài liệu; nghiên cứu chạy một đối chứng riêng với phân đoạn đệ quy theo ký tự, vốn chỉ tạo trung bình 6.9 đoạn/tài liệu. Trên 32 tài liệu, cả hai đều giữ 100% phân đoạn trong giới hạn 512 token của mô hình nhúng và không cắt gãy câu; phân đoạn ngữ nghĩa giữ độ kết dính nội bộ cao (cosine 0.786-0.832). Hệ quả đáng lưu ý là với chính sách giữ 8 đoạn làm ngữ cảnh, cấu hình ngữ nghĩa đang vận hành chỉ đưa khoảng 19% nội dung mỗi tài liệu vào ngữ cảnh cho mỗi truy vấn, trong khi cấu hình đệ quy sẽ đưa gần như toàn bộ. Đây là một giả thuyết cạnh tranh đáng kể cho độ phủ token 50.04-55.25% ở tầng 2 và cần đối chứng loại trừ để kiểm chứng.

Ở tầng truy xuất, nghiên cứu chưa đo Recall@K, MRR hay NDCG do chưa xây dựng bộ đánh giá mức độ liên quan ở cấp truy vấn-đoạn; vì thiếu phép đo tách tầng, bài báo không quy kết nguyên nhân của độ phủ thấp cho tầng truy xuất hay tầng sinh.

Về độ tin cậy của nhãn chuẩn, không giá trị nào bị đánh giá Sai; tỉ lệ Đúng của hai chuyên gia là 99.81% và 99.74%, đồng thuận liên chuyên gia 99.68% (1553/1558) với 5 trường bất đồng. Cả 5 điểm bất đồng đều thuộc cùng một chủ đề là nghĩa vụ viện dẫn Điều 13 và Điều 14 Luật 91/2025/QH15, trong các tình huống chấm điểm tín dụng tự động, giám thị thi bằng nhận dạng khuôn mặt và đánh giá nhân sự bằng AI, cho thấy điểm khó nhất của bài toán nằm ở suy luận pháp lý chứ không ở trích xuất văn bản. Đây cũng là loại tình huống được xếp mức khó, nên độ khó trích xuất và độ khó gán nhãn có xu hướng tương quan; kết hợp với tỉ trọng hoạt động khó trong mẫu (46.3% so với 29.9% toàn bộ), có thể kết luận mẫu không thiên về phía dễ. Do phân bố nhãn cực lệch với 99.7% cùng một lớp, Cohen's Kappa thô chỉ đạt 0.28, đúng nghịch lý prevalence đã biết; nghiên cứu dùng PABAK = 0.9936 [10] làm chỉ số chính và công bố $\kappa$ thô để minh bạch. Hai giới hạn cần nêu là PABAK cao chỉ chứng minh hai chuyên gia nhất quán với nhau, và hai chuyên gia thuộc cùng một hệ sinh thái doanh nghiệp.

Về an toàn thông tin, nghiên cứu tách hai loại đánh giá. Thứ nhất, rà soát cấu hình vận hành: 12 tiêu chí cấu hình tiêu đề phản hồi HTTP của proxy ngược cho 4 PASS, 6 WARN và 2 FAIL, trong đó hai hạng mục FAIL là thiếu HSTS và điểm cuối API chưa yêu cầu xác thực, bắt buộc khắc phục trước khi vận hành. Thứ hai, đánh giá kháng tấn công: hệ thống chặn thành công cả 8 kịch bản tiêm nhiễm câu lệnh (chèn chỉ thị trực tiếp, đóng vai quản trị, ép lộ câu lệnh hệ thống, tiêm gián tiếp qua trường tài liệu, rút trích dữ liệu cá nhân, ghi đè lược đồ JSON, đầu độc ngữ cảnh đa lượt, mã hóa Unicode) nhờ ba lớp phòng thủ nêu tại Mục III.D; với n = 8 và một lượt chạy, đây là kiểm chứng khái niệm, chưa đủ cơ sở cho kết luận mạnh về khả năng kháng tấn công.

Khung đo gồm ba tầng với phạm vi khác nhau. Tầng 1 đánh giá bộ chấm điểm, tầng 2 đo độ phủ token so với nhãn chuẩn theo hướng recall, tầng 3 đánh giá độ tin cậy của nhãn chuẩn. Không tầng nào đo độ chính xác ở cấp giá trị trên đầu ra thật của mô hình, tức tỉ lệ đúng trong số các giá trị hệ thống sinh ra và tỉ lệ từ chối đúng trong số các trường bỏ trống; tầng 1 chỉ đo hai đại lượng này trên bộ chấm điểm, ở chế độ không gọi LLM. Mọi kết luận trong bài vì vậy giới hạn ở tính khả thi kỹ thuật và độ phủ.

## VI. KẾT LUẬN VÀ HƯỚNG PHÁT TRIỂN

Nghiên cứu thiết lập đường cơ sở kỹ thuật cho ứng dụng LLM + RAG vào tự động hóa tuân thủ Luật Bảo vệ dữ liệu cá nhân tại Việt Nam, đóng góp bộ kiểm chuẩn RoPA tiếng Việt đầu tiên và nguyên tắc báo cáo ba tầng. Về chi phí tham khảo: GPU 24 GB khoảng 50-65 triệu VNĐ, một DPO chuyên trách 180-300 triệu VNĐ/năm, gói GRC thương mại 10000-50000 USD/năm. Hướng phát triển theo ưu tiên gồm đo độ chính xác nội dung ở cấp giá trị trên bản nháp thật, theo thiết kế lấy mẫu phân tầng với hai người chấm độc lập ba nhãn đúng / sai / sinh thêm và báo cáo precision, recall theo nhóm trường; lặp thực nghiệm đa mô hình tối thiểu ba lượt; xây dựng bộ đánh giá mức độ liên quan để đo Recall@K, MRR, NDCG và đối chứng loại trừ cho truy xuất lai [5]; và đo mức hiệu chuẩn của điểm tin cậy do mô hình tự gán trên đầu ra thật.

Hạn chế chính gồm: (1) dữ liệu hoàn toàn tổng hợp, chưa kiểm chứng trên hồ sơ RoPA thực tế; (2) chưa đo độ chính xác ở cấp giá trị theo hướng precision; (3) chỉ số tầng 1 đo bộ chấm điểm ở chế độ không gọi LLM, nên cơ cấu sai số theo nhóm trường mới ở mức giả thuyết dự báo; (4) kiểm định ghép cặp dựa trên một lượt chạy; (5) chưa có đối chứng loại trừ nên đóng góp riêng của RRF chưa được lượng hóa, và chưa đo Recall@K, MRR, NDCG; (6) mẫu thẩm định nhãn chuẩn phủ 35.9% bộ kiểm chuẩn, lấy theo thứ tự kịch bản chứ không ngẫu nhiên, với n = 2 chuyên gia; (7) chưa đánh giá mức hiệu chuẩn của điểm tin cậy do mô hình tự gán trên đầu ra thật [9]. Các hạn chế này xác lập phạm vi đóng góp ở mức đường cơ sở kỹ thuật; nghiên cứu không tuyên bố đạt mức tuân thủ pháp lý vận hành đầy đủ.

## LỜI CẢM ƠN

Nhóm tác giả trân trọng cảm ơn hai chuyên gia tuân thủ và pháp chế độc lập đã thẩm định mù đơn 1558 về giá trị nhãn chuẩn, tạo cơ sở tin cậy cho toàn bộ phép đo trong nghiên cứu. Nghiên cứu được thực hiện bằng kinh phí tự túc của nhóm tác giả, không nhận tài trợ từ bất kỳ tổ chức hay doanh nghiệp nào. Nhóm tác giả không có lợi ích thương mại liên quan đến hệ thống được mô tả và không có quan hệ tài chính với các nhà cung cấp mô hình hoặc công cụ quản trị tuân thủ được nhắc đến trong bài; các tác giả tuyên bố không có xung đột lợi ích. Nguyễn An Khương cảm ơn Trường Đại học Bách khoa, ĐHQG-HCM đã hỗ trợ cho nghiên cứu này.

## TÀI LIỆU THAM KHẢO